\documentclass[preprint,12pt]{elsarticle}
\usepackage{xcolor}

\usepackage{amssymb}
\usepackage{amsmath}
\usepackage{url}
\usepackage{wrapfig} 
\usepackage{algorithm}
\usepackage{algorithmic}
\usepackage{booktabs}
\usepackage{multirow} 
\usepackage{caption}
\usepackage{multirow}
\usepackage{makecell}
\usepackage{multirow}
\usepackage{algorithm}
\usepackage{algorithmic}
\usepackage{makecell}
\usepackage{graphicx}
\usepackage{subcaption}  
\usepackage{float}  
\usepackage{caption}  
\usepackage{wrapfig} 
\usepackage{times}
\usepackage{url}

\usepackage{lineno}

\begin{document}

\begin{frontmatter}



\title{Domain Generalizable Knowledge Tracing via Concept Aggregation and Relation-Aware Attention}



\author{Yuquan Xie, Shengtao Peng, Wanqi Yang, Ming Yang, Yang Gao}

\begin{abstract}

Knowledge Tracing (KT) is a fundamental task in online education systems, aiming to model students’ knowledge states by predicting their future performance. However, in newly established educational platforms, traditional KT models often fail due to the scarcity of student interaction data. To address this challenge, we leverage interaction data from existing education systems. Yet, these data sources differ substantially in subjects, exercises, and knowledge concepts.

We formulate this problem as a multi-source domain adaptation scenario for cold-start knowledge tracing and propose a novel domain-generalizable framework, DGKT. Our approach constructs generalized concept prototypes to unify semantic representations across domains and reduces domain discrepancies through a Sequence Instance Normalization (SeqIN) strategy. To better capture temporal dependencies in student behavior, we further introduce a Relation-Aware Attention Encoder (RA-Encoder).
We evaluate our method on five benchmark datasets with extremely limited data, where each target domain contains fewer than 300 labeled training samples. Experimental results show that DGKT framework achieves an average AUC improvement of 1.98\% across domains. Furthermore, our proposed DGrKT model outperforms existing KT models with a notable 4.16\% AUC improvement.
The source code is publicly available at: \url{https://anonymous.4open.science/r/DGKT-5330}.

\end{abstract}


\end{frontmatter}



\section{Introduction}

Over the past decades, the emergence of numerous online education systems has significantly transformed the educational landscape by offering remote learning environments and personalized guidance for users. These advancements have revolutionized the educational landscape, enabling platforms to cater to the individual learning needs of students more effectively \cite{anderson2014engaging, lan2014time, zhao2018automatically}. In this context, knowledge tracing (KT) plays a crucial role as it allows education systems to monitor and evaluate the evolving knowledge states of students. Knowledge tracing (KT) specifically aims to accurately track a student's knowledge progression, typically through the prediction of future academic performance based on their historical interactions~\cite{abdelrahman2023knowledge, baker2009state, yu2025exploring}. This enables online education platforms to better assess students' comprehension and provide tailored assistance for learners~\cite{piech2015deep,song2022bi}.

\begin{wrapfigure}{r}{0.6\textwidth}
    \centering
    \includegraphics[width=\linewidth]{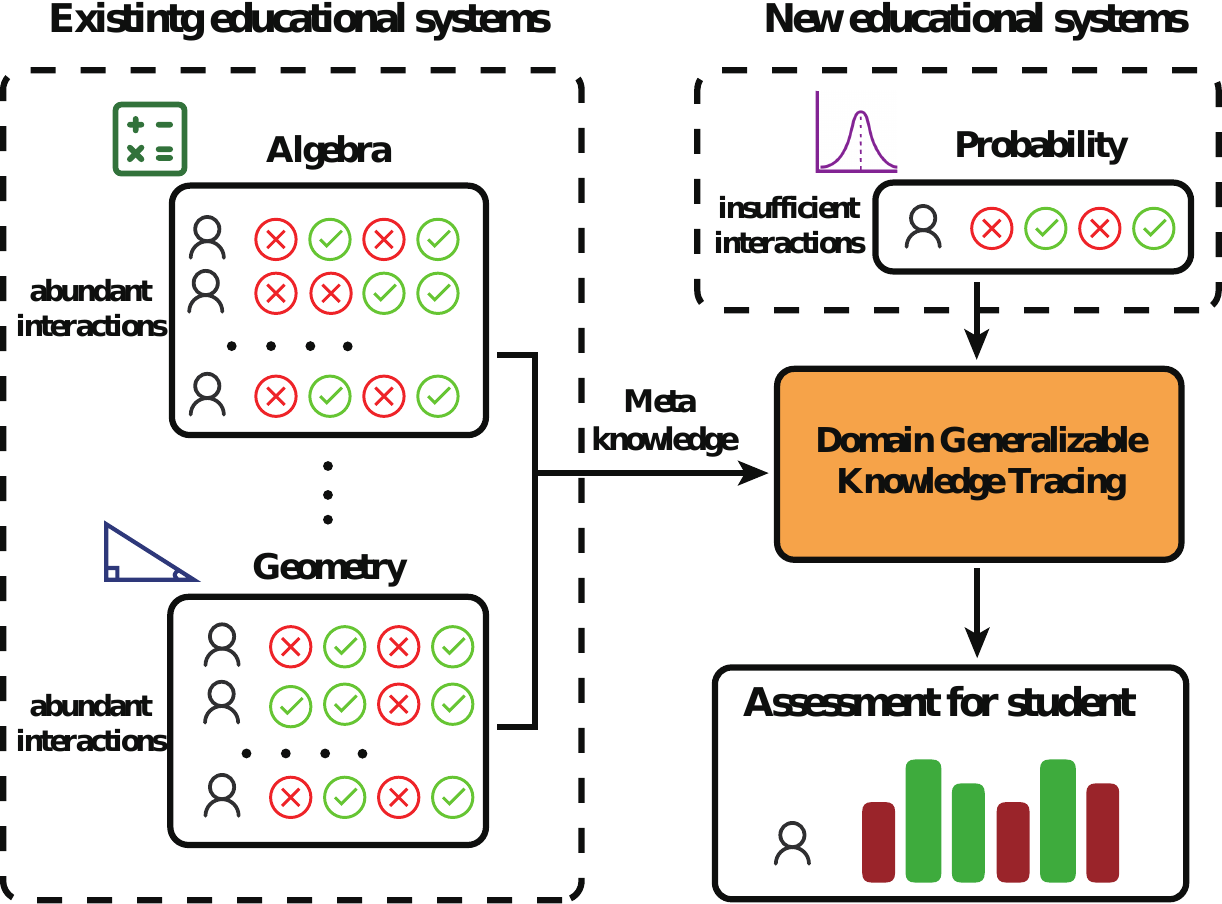}
\caption{An intuitive explanation for DGKT. Existing systems contain abundant interaction records while new education system contains only a few interactions. Our goal is to first train a versatile knowledge tracing model to acquire meta-knowledge from existing education systems and then apply it to the new system.}
        \label{intro}
\end{wrapfigure}

\begin{figure}
    \centering
    \includegraphics[width=0.7\columnwidth]{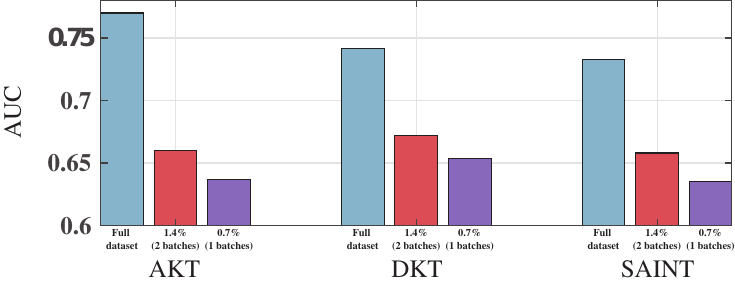}
  \caption{The AUC result for three existing KT models (DKT \cite{piech2015deep}, AKT \cite{ghosh2020context}, SANIT \cite{choi2020towards}) on ASSISTment 2009 with different data sizes where the proportion of available data gradually decreases.}
  \label{datasize}
\end{figure}

A number of KT models have demonstrated their effectiveness \cite{choi2020towards,ghosh2020context,DBLP:journals/tkde/AbdelrahmanW23, jiang2024improving, ma2024hd, sun2025daskt}. However, these models typically require a large volume of student interactions to sufficiently train a KT model for a particular subject area. In the majority of knowledge tracing datasets, students' problem-solving records typically have a scale of several hundred thousand or more. For example, there are 330, 116 interactions in ASSISTment 2009 to train a KT model sufficiently while \textit{it is challenging for newly developed education systems to accumulate such a substantial volume of data at the outset. }
In practical scenarios, the development of a new online education system or the introduction of a new question bank within an existing platform often lacks the extensive data on student interactions needed for effective KT model training \cite{liu2012cocktail, wilson2016estimating}. This scarcity of interactions results in the challenge of insufficient training data. As illustrated in Fig. \ref{datasize}, common KT models suffer a notable AUC degradation when confronted with limited data sizes.

\textbf{In cases where student interaction records are scarce in the new system, numerous student interaction records from various online education platforms provide valuable insights.} These records may originate from different sources, yet they mirror the process of knowledge acquisition. 
Extracting overarching cognitive patterns from these varied student interaction sequences is beneficial for knowledge tracing in new systems.
However, interaction sequences from different education systems differ a lot, so it is unfeasible to directly use them as training data.

Thus, we innovatively formulate the dilemma of the new education system as a Multi-source Domain Adaptation setting for cold-start KT, as shown in Fig. \ref{intro}. Sufficient student interaction records from various online education domains, such as Algebra and Geometry, are treated as source domains. Meanwhile, the limited interactions from the new system, such as those in Probability, are treated as the target domain.
\textit{Our goal is to train a domain-generalizable model that extracts meta-knowledge from the diverse student interactions across source domains, allowing for a seamless transfer of this knowledge to the target domain.}.

\begin{wrapfigure}{r}{0.5\textwidth}
    \centering
    \includegraphics[width=\linewidth]{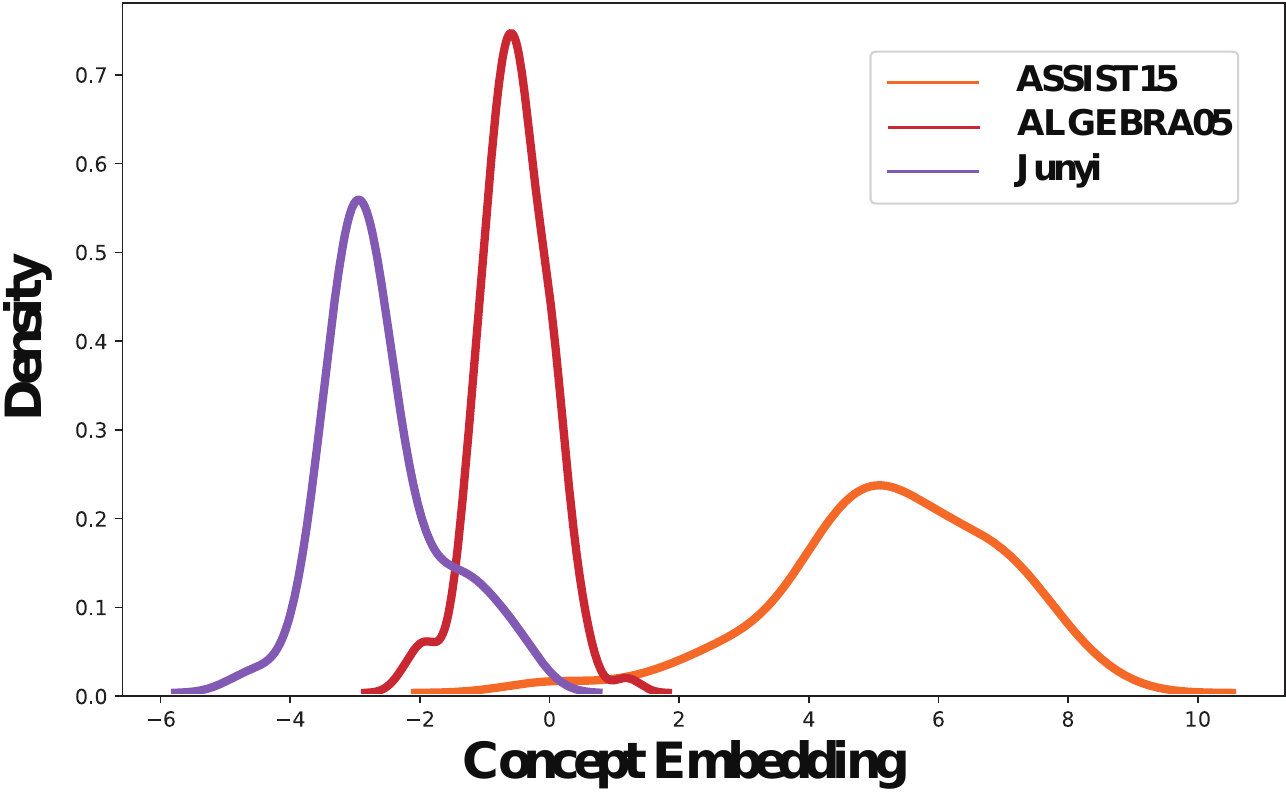}
\caption{Kernel Density Estimate (KDE) for concept embedding from different education systems, \emph{i.e.,} ASSIST15, ALGEBRA05, and Junyi.}
        \label{kde}
\end{wrapfigure}

To realize this goal, we propose a novel Domain Generalizable Knowledge Tracing (DGKT) framework.
Notably, DGKT faces two primary challenges: \textit{1) significant discrepancy among the source domains and 2) scarce student interactions within the target domain.}

\textit{For the problem of significant distribution discrepancy among the source domains}, as clearly illustrated in Fig. \ref{kde}, we perform a kernel density estimation for the knowledge concepts from various domains, which reveals significantly distinct distributions for the concept embeddings from different domains. 
To overcome this challenge, we propose the concept aggregation—an algorithm that aggregates concept embeddings from different domains into concept prototypes for subsequent analysis of students’ interaction sequences.

\textit{For the problem of scarce student interactions within the target domain}, we recognize the challenge of learning accurate embeddings of the target question with limited interaction data. Consequently, we design a unique concept representation for the target domain that can effectively adapt to target domain with a few training data. Moreover, we propose a Relation-Aware Attention Encoder (RA-Encoder) that fully leverage the relation of the exercises in target domain. The contributions of this paper are summarized as follows:
\begin{itemize}
\item \textbf{A novel domain-generalizable framework for cold-start KT.}
This issue is valuable but insufficiently studied. We formulate the multi-source domain adaptation setting and propose the DGKT framework that can quickly adapt to the target domain with minimal data. The DGKT framework can be integrated into existing KT models.

\item \textbf{Innovative concept aggregation for reducing domain discrepancy.}
This addresses the significant challenge posed by the vast differences in concepts across various domains. Through concept aggregation, similar concepts are aggregated into a concept prototype.

\item \textbf{The Relation-Aware Knowledge Encoder with better generalizability.}
We propose a relation-aware knowledge encoder (RA-Encoder) which fully leverages relational information within the target domain. By integrating RA-Encoder into DGKT framework, we construct the Domain-Generalizable Relation-aware Knowledge Tracing model (DGrKT). Our specially designed model demonstrates superior performance in multi-source domain adaptation for cold-start knowledge tracing.
\end{itemize}

After conducting extensive experiments on five benchmark datasets, our DGKT framework achieves an average AUC improvement of 2.13\%. Moreover, our DGrKT demonstrates the best performance, with an AUC improvement of 4.16\% over five existing KT methods on multi-source domain adaptation for cold-start knowledge tracing tasks.

\section{Related Work}
Here, we introduce the related work including knowledge tracing, cold start and multi-source domain adaptation.

\subsection{Knowledge Tracing}
Knowledge tracing task is an essential task to trace a student's knowledge state over an extended learning period. Previous methods for KT can be divided into two categories, \emph{i.e.,} traditional machine learning methods and deep learning methods. Among the methods based on traditional machine learning algorithms, the most representative one is BKT~\cite{corbett1994knowledge}. BKT builds a hidden Markov model for each knowledge concept to predict a student's mastery of specific concepts.
Other traditional machine learning KT models include Performance Factors Analysis (PFA)~\cite{pavlik2009performance} and item response theory (IRT)~\cite{ebbinghaus2013memory}.

Recently, many deep models for KT have emerged. The earliest deep model for KT is Deep Knowledge Tracing (DKT)~\cite{piech2015deep}. DKT applies recurrent neural networks (RNNs) and outperforms traditional KT models. Many variants of DKT are proposed afterwards such as DKT+~\cite{yeung2018addressing}. Exercise-Enhanced Recurrent Neural Network (EERNN)~\cite{su2018exercise} is proposed for student performance prediction by taking full advantage of both student exercising records and the text of each exercise. Another typical deep KT model is Separated Self-AttentIve Neural Knowledge Tracing (SAINT)~\cite{choi2020towards} which applies Transformer to KT tasks. Attentive Knowledge Tracing (AKT)~\cite{ghosh2020context} is another attention-based KT model using a novel monotonic attention mechanism that relates a learner’s future responses to assessment questions to their past responses. Learning Process-consistent Knowledge Tracing (LPKT)~\cite{shen2021learning} monitors students' knowledge state through directly modeling their learning process. DASKT~\cite{sun2025daskt} incorporates simulated affective states into KT, improving interpretability and performance. LefoKT \cite{bai2025rethinking} introduces relative forgetting attention to decouple forgetting from item relevance and improve attention-based KT models’ ability to handle ever-growing interaction sequences. AdaptKT~\cite{cheng2022adaptkt}, delved into domain adaptation for knowledge tracing with text information and relatively sufficient data. As far as we know, few works focus on the data scarcity issue in online education systems, and we innovatively attempt to train a generalizable KT model.

Currently, few studies have delved into the data scarcity issue in knowledge tracing. Some prior efforts have aimed to address the sparse problem that students tend to interact with only a small set of questions, using approaches like pre-trained question embeddings~\cite{liu2020improving} or contrastive learning~\cite{song2022bi, yang2023contextualized, yang2021learning}. 
In cases where interactions are severely limited, these approaches might not be effective in learning question embeddings and students' knowledge states, as they heavily rely on a larger amount of student interactions. \textit{The scarcity issue of student interactions in new educational systems presents a significant challenge for knowledge tracing.}

\subsection{Cold Start and Multi-Source Domain Adaptation}
The cold-start problem, where data is scarce, is a significant challenge in various real-world applications such as recommender systems and personalized learning. One promising solution to this issue is source-free domain adaptation (SFDA), which leverages knowledge from source domains with abundant data to assist target domains with limited data. SHOT \cite{ahmed2021unsupervised} introduces a novel method to address this problem by employing a clustering-based pseudo-labeling technique and incorporating information maximization loss to align target domain features with the pretrained source model. In a similar vein, \cite{kundu2022balancing} highlight the trade-offs between discriminability and transferability and propose a strategy that combines original and translated samples using mix-up to enhance model performance.

When data from multiple source domains is available, Multi-source Domain Adaptation (MSDA) becomes a more effective approach, enabling knowledge transfer from various domains. In this context, \cite{ahmed2021unsupervised} expands the SFDA framework by combining the outputs from different source models with learnable weights, and adapting the models using weighted information maximization. CAiDA \cite{dong2021confident} builds upon this idea by integrating a pseudo-label generator based on confident-anchor induction, further improving the adaptation process. DATE \cite{han2023discriminability} takes a different approach by evaluating the transferability of source models through a Bayesian framework, quantifying the similarity between domains via a multi-layer perceptron. Bi-ATEN \cite{li2024agile} introduces a tuning-free bi-level attention ensemble to adapt multiple source models to an unlabeled target domain without source data.

\section{Domain-Generalizable Knowledge Tracing}
In this section, we elaborate on the proposed DGKT framework. Firstly, we formulate the KT tasks and their multi-source domain adaptation setting. Subsequently, we present the DGKT architecture. Furthermore, we introduce the process of concept aggregation. Finally, we introduce the RA-Encoder, which is specifically designed to address the multi-source domain adaptation problem.

\subsection{Problem Definition}
\textbf{A KT task with few sequences.} In a KT task, there are very few sequences of interactions for training, denoted as $I = \{(q_1, r_1), (q_2, r_2), ..., (q_T, r_T)\}$. Here, $T$ represents the length of the sequence, $q_t \in \mathbb{N}^+$ corresponds to the question ID of the $t$-th interaction ($t\leq T$), and $r_t \in \{0, 1\}$ means the correctness of the student's answer to the question $q_t$. 
These questions are associated with $n_c$ concepts that students need to master. The objective is to train a KT model capable of mining the knowledge state of students and predicting the probability that a student will answer the next question correctly, denoted as $P(r_{n+1} | q_{n+1},I)$. Since data scarcity affects model training, we resort to other available sequences as auxiliary source domain data for multi-source domain adaptation.

\textbf{Multi-source domain adaptation setting for cold-start KT.} In the context of multi-source domain adaptation for cold-start knowledge tracing, there are $N$ auxiliary source domains $\{D_s^i | 1\leq i \leq N\}$ and a target domain $D_t$. Each source domain $D_s^i$ consists of $m_i$ sequences of student interactions, denoted as $D_s^i = \{I_j^i | 1\leq j \leq m_i\}$ along with a set of questions $\{q_1, q_2,...,q_{n_{q_i}}\}$ and a set of concepts $\{c_1, c_2,...,c_{n_{c_i}}\}$, where $n_{q_i}$ represents the number of questions, $n_{c_i}$ represents the number of concepts. The target domain, on the other hand, contains only $m_t$ student interaction sequences, represented as $D_t = \{I_j^t | 1 \leq j \leq m_t\}$, with $m_t \ll m_i$. \textit{Importantly, the data $D_t$ from the target domain is unseen during the meta-training phase.} The objective is to train a generalized KT model by leveraging all the source domain data $\{D_s^i|1 \leq i \leq N\}$, and subsequently adapt this model to the target domain data $D_t$ for knowledge tracing.

\begin{figure*}
  \centering
  \includegraphics[width=\textwidth]{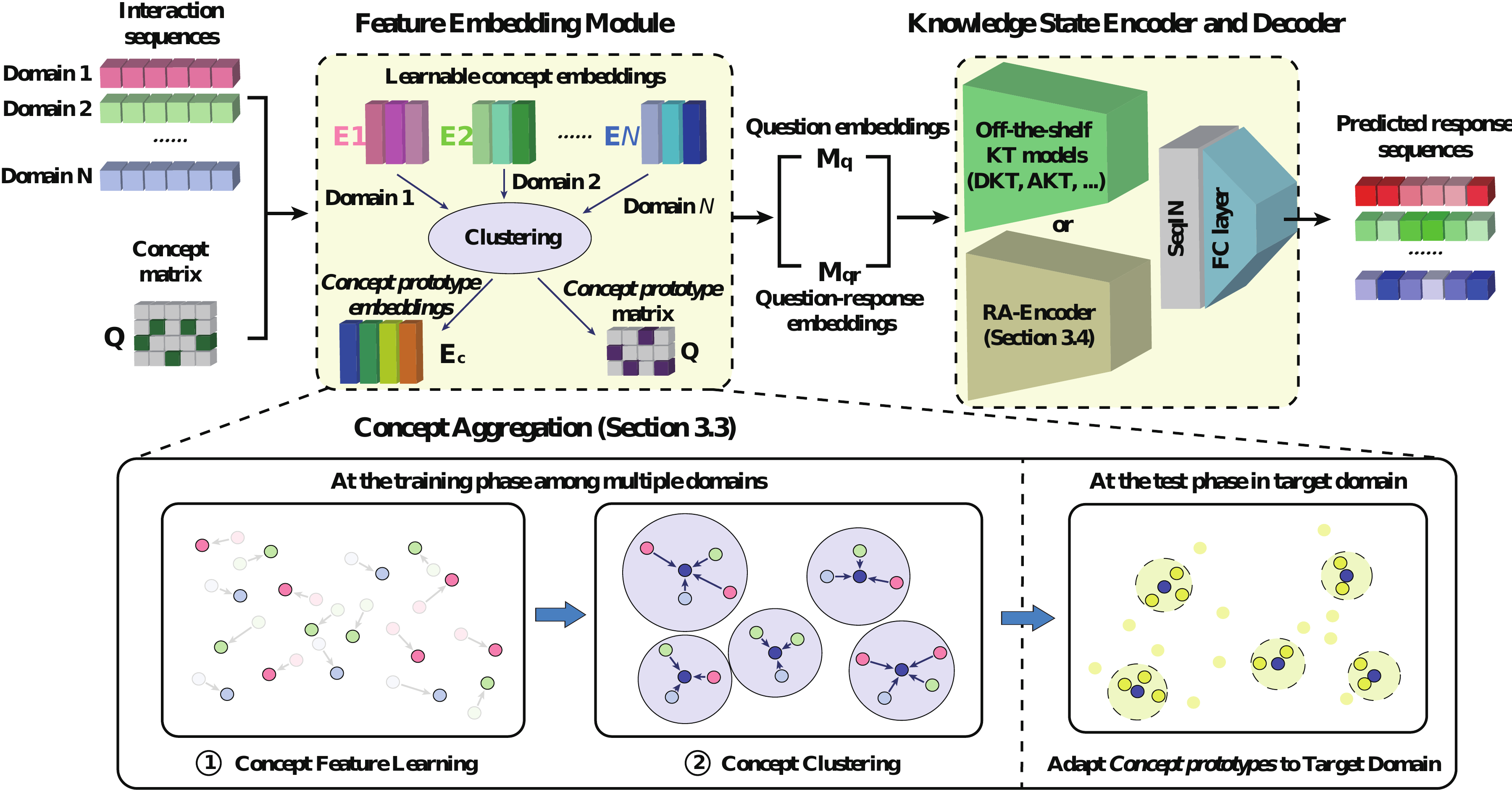}
  \caption{A concise overview of our DGKT. We employ concept aggregation to obtain the embeddings that are applicable across domains. We conduct concept feature learning first and then aggregate embeddings into several clusters to get a refined centroid embbedding. Finally, we utilize learned centroid embedding to represent target concepts.}\label{framework}
\end{figure*}

\subsection{DGKT Architecture}\label{modelarchitecture}
The process outlined in Fig. \ref{framework} illustrates the key steps of our DGKT approach. Initially, we employ the \textit{feature embedding module} to transform the interaction sequences into corresponding embedding sequences, followed by the \textit{knowledge state encoder} that generates the hidden knowledge states. Then, the knowledge states are decoded by the \textit{knowledge state decoder}, finally producing the anticipated probability of a student providing a correct response to the subsequent question.

\textbf{Feature embedding module.}
In feature embedding module, our objective is to convert the sequence of questions and responses into the embedding sequence.
For each domain, we have a concept matrix $\textbf{Q} \in \mathbb{R}^{n_c \times n_q}$, where $n_q$ represents the total number of question IDs and $n_c$ represents the total number of knowledge concept IDs. The element $\textbf{Q}_{ij}$ is set to 1 if the $j$-th question is related to the $i$-th knowledge concept. 
Then, we learn the embedding $e_{q_t}\in \mathbb{R}^d$ of question $q_t$ ($d$ is the dimension of the embedding), by averaging the embeddings $e_i$ of all concepts associated with the question, which can be written by

\begin{equation}\label{embedding_source}
e_{q_t} = \frac{\sum_{i} \mathbb{I}(\textbf{Q}_{iq_t} = 1) e_i}{\sum_{i} \mathbb{I}(\textbf{Q}_{iq_t} = 1)},
\end{equation}
where $\mathbb{I}(\cdot)$ represents the indicator function, and $e_i\in\mathbb{R}^{d}$ is a learnable vector representing the $i$-th concept.

Afterward, we leverage the response $r_t$ of the question to construct the question-response embedding $e_{qr_t}\in \mathbb{R}^{2d}$ as follows:
\begin{equation}\label{embedding_source2}
e_{qr_t} = \begin{cases}
e_{q_t} \oplus \boldsymbol{0}, & \text{if } r_t = 1,\\
\boldsymbol{0} \oplus e_{q_t}, & \text{if } r_t = 0,
\end{cases}
\end{equation}
where $\boldsymbol {0} = (0,0,...,0)\in \mathbb{R}^{d}$ is an all zero vector with the same dimension $d$ as $e_{q_t}$, and $\oplus$ is the concatenation operator.

By concatenating all the embeddings in a sequence, we obtain the question embedding matrix $\textbf{M}_{q\{1:T\}} = (e_{q_1}, e_{q_2}, ..., e_{q_T}) \in \mathbb{R}^{d \times T}$ and the question-response embedding matrix $\textbf{M}_{qr\{1:T\}} = (e_{{qr}_1}, e_{{qr}_2}, ..., e_{{qr}_T}) \in \mathbb{R}^{2d \times T}$.

\textbf{Knowledge state encoder using existing models.} 
To encode the question embeddings and question-response embeddings into knowledge state, many off-the-shelf knowledge tracing models can be used as the knowledge state encoder: 
\begin{equation}
    \{h_t\} = \text{Enc}(\textbf{M}_q, \textbf{M}_{qr}),
\end{equation}
where the embedding matrices $\textbf{M}_q$ and $\textbf{M}_{qr}$ are encoded into obtain knowledge state $\{h_t\}$ at different timestamps.

Here we take DKT \cite{piech2015deep}, SAINT \cite{choi2020towards} and AKT \cite{ghosh2020context} as examples to illustrate how to transform embedding into knowledge state. Note that the architecture of SimpleKT \cite{liu2023simplekt} can also be used, but we will not elaborate on it due to its simplicity and similarity to SAINT and AKT.

For DKT \cite{piech2015deep}, we use LSTM as knowledge state encoder to get student's knowledge state:

\begin{equation}
    h_t = \text{LSTM}(e_{qr_t}, h_{t-1}),
\end{equation}
where $e_{qr_t}$ denotes the $t$-th question-response embedding and, $h_t$ represents the student's knowledge state at timestep $t$. The LSTM network takes knowledge state $h$ as its hidden state.

As for SAINT \cite{choi2020towards}, it usually utilizes a Transformer architecture. The Transformer's encoder is responsible for receiving the student's question embedding matrix, while its decoder receives the student's question-response embedding matrix along with the output from the encoder, which can be written by:
\begin{equation}
\begin{split}
    \textbf{O}_{\{1:t\}} &= \text{Encoder}(\textbf{M}_{q\{1:t\}}), \\
    h_t &= \text{Decoder}(\textbf{M}_{qr\{1:t-1\}}, \textbf{O}_{\{1:t\}}),
\end{split}
\end{equation}
where 
Encoder and Decoder represent the Transformer's encoder and decoder, respectively.

Likewise, in the AKT \cite{ghosh2020context} model, three attention-based modules are utilized, which are the question encoder, the knowledge encoder, and the knowledge retriever. The student's knowledge state can be represented as:
\begin{equation}
\begin{split}
    \textbf{X}_{\{1:t\}} &= \text{Encoder}_q(\textbf{M}_{q\{1:t\}}), \\
    \textbf{Y}_{\{1:t-1\}} &= \text{Encoder}_k(\textbf{M}_{qr\{1:t-1\}}), \\
    h_t &= \text{Decoder}(\textbf{X}_{\{1:t\}}, \textbf{Y}_{\{1:t-1\}}), 
\end{split}
\end{equation}
where $\text{Encoder}_q$ represents the question encoder which produce contextualized representations of each question. $\text{Encoder}_k$ represents the knowledge encoder which produces contextualized representations of each  question and response. Decoder represents the knowledge retriever which retrieves knowledge state.

In sum, the knowledge state encoder generates the knowledge state $h_t$, which will be sent through the knowledge state decoder.

\textbf{Knowledge state decoder.}
We use a unified knowledge decoder to predict the probability of a student answering the next question correctly:
\begin{equation}
\begin{aligned}
\label{decoder}
    \hat{y}_{t+1}  &= \text{Dec}(h_{t+1}, q_{t+1}) \\
    &= \sigma (W_2 \cdot \text{ReLU}(W_1 \cdot [h_{t+1}, e_{q_{t+1}}] + b_1) + b_2),
\end{aligned}
\end{equation}
where Dec represents the knowledge state decoder, and $W_1$, $W_2$ and $b_1$, $b_2$ denote the weights and the biases, respectively. Also, $\sigma(\cdot)$ is the sigmoid function.

\textbf{Sequence Instance Normalization}
To reduce the distribution discrepancy from different domains, utilizing normalization method is a common solution. However, existing normalization methods (\emph{e.g.,} Batch Normalization \cite{bn} and Instance Normalization \cite{in}) may not be applicable for sequential features in knowledge tracing tasks. For example, Instance Normalization normalizes all the information within a single channel, which can lead to the information leakage while predicting student's performance at timestep $t$. This is because the student's interactions after timestep $t$ should remain unseen in KT task. To tackle the dilemma, we design the Sequence Instance Normalization (SeqIN) method in our KT model to normalize the feature embeddings of sequential student interactions among domains. Since the normalization process takes into account the fact that the later interactions cannot be seen in the previous interactions, their feature embeddings at the current moment are normalized by only using the statistics of all the previous moments.

Given a sequential feature embedding matrix $\textbf{M} = (m_1, m_2, ..., m_n)$ where $m_t \in  \mathbb{R}^d$ is the feature embedding at timestep $t$, we firstly calculate the mean $\mu_t$ and standard deviation $\sigma_t$ before timestep $t$:

\begin{equation}
\mu_t = \frac{1}{t+1} (p + \sum_{i=1}^t m_i),
\end{equation}

\begin{equation}
\sigma_t = \sqrt{\frac{(p- \mu_t)^2 + \sum_{i=1}^t (m_i - \mu_t)^2}{t+1}},
\end{equation}
where 
$\mu_t$ and $\sigma_t$ denote the mean and standard deviation of the sequence $\{m_1,m_2,...,m_t\}$, and $p$ is a learnable padding vector. In consideration of the meaninglessness to compute the mean and standard deviation of $m_1$ itself, we add a padding vector $p$ in front of the original sequence, \emph{i.e.,} $\{p,m_1,m_2,...,m_n\}$, where $p$ is learned along with the model.

We then normalize the sequential feature embedding matrix $\textbf{M}$ into $\Tilde{\textbf{M}}$ using the calculated mean and standard deviation:

\begin{equation}
    \Tilde{m}_t = \gamma(\frac{m_t-\mu_t}{\sigma_t})+ \beta,
\end{equation}

\begin{equation}
    \Tilde{\textbf{M}}=[\Tilde{m}_1, \Tilde{m}_2, ...,  \Tilde{m}_n],
\end{equation}
where $\gamma, \beta \in \mathbb{R}^d$ are the learnable parameters. As seen, SeqIN normalizes $m_t$ by considering all previous feature embeddings up to time $t$ while remaining later feature embeddings unseen. 

We apply SeqIN to aforementioned knowledge state encoders, including DKT, SAINT and AKT.
For DKT, the SeqIN is directly used on students' knowledge state $h$:
\begin{equation}
    \Tilde{h} = SeqIN(h),
\end{equation}
where $\Tilde{h}$ is the normalized students' knowledge state.

As for SAINT, we apply SeqIN to intermediate features $o$:
\begin{equation}
    \Tilde{o} = SeqIN(o),
\end{equation}
where $o$ is the output of Transformer's encoder.

For AKT, we apply SeqIN to intermediate features $x$ and $y$ as follows:
\begin{align}
    \Tilde{y} = SeqIN(y), \\
    \Tilde{x} = SeqIN(x),
\end{align}
where $x$ and $y$ is respectively derived from the question encoder and the knowledge encoder of AKT.

It is worthy noting that SeqIN can effectively aggregate the feature embeddings of student interactions from different source domains to reduce the domain gaps, which can be observed in the visualization experiment in Figure \ref{tsne}.

\subsection{Concept Aggregation}\label{concept_aggregation}
To overcome significant distribution discrepancy among the source domains, we design the concept aggregation procedure for the KT model, which enables the retrieval of meta-knowledge from source domains through aggregating the concepts from diverse source domains, subsequently allowing for adaptation to the target domain.

In the knowledge tracing task, each question is typically associated with a few specific knowledge concepts, which is directly utilized by conventional knowledge tracing models. However, this approach is not suitable for multi-source domain adaptation knowledge tracing tasks, as knowledge concepts can differ significantly across various domains. On one hand, the feature embeddings of concepts from different domains exhibit significant distribution variations, as they belong to different education systems. On the other hand, these concepts may exhibit shared attributes, including difficulty, forgettability, among others.

To mitigate the challenge of substantial variations among concepts while maintaining their shared characteristics, we propose a concept aggregation strategy to generate \emph{concept prototypes}. These prototypes represent the concepts from a unified perspective by aggregating similar domain-specific knowledge concepts. This concept aggregation process involves: 1) \textbf{Concept feature learning} for domain-specific concept embeddings, 2) \textbf{Concept clustering} by conducting $k$-means algorithm for the concept embeddings across all source domains, and subsequently, 3) \textbf{Application of \emph{concept prototypes}} for re-representing question embeddings and model updating.

\textbf{Concept feature learning.} Initially, we train the KT model across all source domains. Each source domain generates its own concept embedding matrix, denoted as $\textbf{E}_i = [e_1, e_2, ..., e_{n_i}]$, where $e_i \in \mathbb{R}^d$ represents the embedding of concept $i$. The remaining parameters of knowledge state encoder and decoder, $\theta_{enc}$ and $\theta_{dec}$, are shared across all domains within the KT model. During this phase, data from each source domain is randomly sampled to ensure that the model does not become biased towards the source domain with larger amounts of data. The model is trained by minimizing the classification loss across all source domains, which can be mathematically expressed as:
\begin{equation}
{ \underset {\{\textbf{E}_i\}_{i=1}^N,\theta_{enc}, \theta_{dec}} 
{ \operatorname {\min} } \, 
\mathbb{E}_{D_s^i\sim D_s,(I^i_j,R^i_j)\sim D_s^i} [L_\text{CE}(\hat{Y^i_j}, R^i_j)]  },
\label{L1}
\end{equation}
where $L_\text{CE}$ is the cross-entropy loss, $I^i_j$ is the $j$-th interaction sequence in the $i$-th source domain, $\hat{Y^i_j}$ is the probability generated by the model, and $R^i_j$ is the ground truth, representing the corresponding response.

\textbf{Concept clustering.} After the initial training on source domains, we acquire the concept embeddings $\{\textbf{E}_i\}_{i=1}^N$ from various domains. In this phase, we apply the $k$-means algorithm to all the concept embeddings, grouping numerous concepts into $k$ \emph{concept prototypes}. This procedure aggregates concepts that share common characteristics into a single \emph{concept prototype}, even if these concepts originate from different domains. This approach helps reduce the distribution discrepancy between domains.

Specifically, through the $k$-means procedure, we first obtain a cluster assignment matrix $\textbf{A}$: 
\begin{align}
    &\textbf{E}_{\text{cat}} = [\textbf{E}_1, \textbf{E}_2, ..., \textbf{E}_N], \\\label{kmeans}
    &\textbf{A} = \text{Kmeans}(\textbf{E}_{\text{cat}}),
\end{align}
where $\textbf{E}_{\text{cat}}$ is the concatenation of all concept embeddings from $N$ source domains. The matrix $\textbf{A} \in \{0, 1\}^{k \times n_{\text{e}}}$ represents the cluster assignment matrix, where $\textbf{A}_{ij} = 1$ indicates that the $j$-th concept from $\textbf{E}_{\text{cat}}$ is assigned to the $i$-th \emph{concept prototype}. Then, we calculate the embedding of a \emph{concept prototype} by averaging all concept embeddings in a cluster, and concate all \emph{concept prototypes} to form a matrix, which can be written by:
\begin{align}
    &e_{c_i} = \frac{1}{|C_i|} \sum_{j=1}^{n_{\text{e}}} \textbf{E}_{\text{cat}_j} \textbf{A}_{ij}, \\\label{kmeans4}
    &\textbf{E}_{c} = [e_{c_1},e_{c_2},...,e_{c_k}],
\end{align}
where $e_{c_i}$ denotes the embedding of $i$-th \emph{concept prototype} and $\textbf{E}_{c} \in \mathbb{R}^{d \times k}$ represents the embedding matrix of \emph{concept prototypes}. Here, $n_{\text{e}}$ represents the total of the concepts, and $|C_i|$ denotes the number of concepts of the $i$-th cluster.

\textbf{\emph{Concept prototypes} refinement.}
To reduce the differences between the source domains, we employ the \emph{concept prototypes} to replace the original domain-specific concepts and to re-represent the question embeddings. Subsequently, these \emph{concept prototypes} are further refined to update the model.

Specifically, for the $s$-th domain, we leverage the clustering assignment matrix $\textbf{A}_s$ of concepts and the concept matrix $\textbf{Q}_s$ to calculate the \emph{concept prototype} matrix $\Tilde{\textbf{Q}}$, \emph{i.e.,}

\begin{equation}
    \Tilde{\textbf{Q}} = \textbf{A}_s \cdot \textbf{Q}_s,
\end{equation}
where $\Tilde{\textbf{Q}}_{ij} = 1$ indicates that the $j$-th question in the $s$-th domain is associated with the $i$-th \emph{concept prototype}.  

Then, the question embedding can be calculated by averaging all the embeddings of associated \emph{concept prototypes}:
\begin{align}
\label{embedding_centroid}
    \Tilde{e_{q_t}} &= \frac{\sum_{i} \mathbb{I}(\Tilde{\textbf{Q}}_{iq_t} = 1) e_{c_i}}{\sum_{i} \mathbb{I}(\Tilde{\textbf{Q}}_{iq_t} = 1)},
\end{align}
where $\Tilde{e_{q_t}}$ is reformulated by using \emph{concept prototypes}, as contrasted with Eq. (\ref{embedding_source}). Subsequently, we update the embedding matrix ($\textbf{E}_{c}$) of the \emph{concept prototypes} along with the model parameters ($\theta_{enc}$ and $\theta_{dec}$) within the KT model by minimizing the cross-entropy loss (as referenced in Eq. (\ref{L1})).

\begin{algorithm} 
    \caption{Concept aggregation}
    \label{alg1}
    
\begin{flushleft}
\textbf{Input}: Interactions from $N$ source domains $\{D_s^i\}_{i=1}^N$ and target domain $D_t$.\\
\textbf{Output}: Model parameters $\theta_{enc}, \theta_{dec}, E_c$ and $E_t$.\\
\end{flushleft}

    \begin{algorithmic}[1]
        \STATE\textit{\textcolor{gray}{\small{$//$ Concept feature learning}}}\\
        \WHILE{not converged}
        \STATE $\mathcal{L}_{kt}\leftarrow0$.
        \FOR{$i \leftarrow 1 ~\text{to}~ N$}
        \STATE $\{I^i, y^i\} \leftarrow \text{randselect}(D_s^i)$.\quad{$//$ \textit{\textcolor{gray}{\small Sample a batch}}}

        \STATE $\{\textbf{M}_q, \textbf{M}_{qr}\} \leftarrow$ solution of Eqs. (\ref{embedding_source}) and (\ref{embedding_source2}).
        \STATE $\hat{y} \leftarrow \text{Dec}(\text{Enc}(\textbf{M}_q, \textbf{M}_{qr}))$.\quad{$//$ \textit{\textcolor{gray}{\small Predict $\hat{y}$}}}
        
        \STATE $\mathcal{L}_{kt} \leftarrow \mathcal{L}_{kt} + \mathcal{L}_{\text{CE}}(\hat{y}, y^i)$.\quad{$//$ \textit{\small \textcolor{gray}{Calculate the loss}}}
        \ENDFOR
        \STATE Update $\{\textbf{E}_i\},\theta_{enc}, \theta_{dec}$ by minimizing $\mathcal{L}_{kt}$.
        \ENDWHILE

        \STATE\textit{\small{$//$ 
        \textcolor{gray}{Concept clustering and concept prototype refinement}}}\\
        \STATE $\{\textbf{A}, \textbf{E}_c\} \leftarrow$ solution of Eqs. (\ref{kmeans})-(\ref{kmeans4}).
        

        \WHILE{not converged}
        \STATE $\mathcal{L}_{kt}\leftarrow0$.
        \FOR{$i \leftarrow 1 ~\text{to}~ N$}
        \STATE $\{I^i, y^i\} \leftarrow \text{randselect}(D_s^i)$.\quad{$//$ \textit{\small \textcolor{gray}{Sample a batch}}}
        \STATE $\{\textbf{M}_q, \textbf{M}_{qr}\} \leftarrow$ 
        solution of Eqs. (\ref{embedding_centroid}) and (\ref{embedding_source2}).
        \STATE $\hat{y} \leftarrow \text{Dec}(\text{Enc}(\textbf{M}_q, \textbf{M}_{qr}))$.\quad{$//$ \textit{\small \textcolor{gray}{Predict $\hat{y}$}}}
        \STATE $\mathcal{L}_{kt} \leftarrow \mathcal{L}_{kt} + \mathcal{L}_{\text{CE}}(\hat{y}, y^i)$.\quad{$//$ \textit{\small \textcolor{gray}{Calculate the loss}}}
        \ENDFOR
        \STATE Update $\textbf{E}_c,\theta_{enc}, \theta_{dec}$ by minimizing $\mathcal{L}_{kt}$.
        \ENDWHILE
        
        \STATE\textit{\textcolor{gray}{\small{$//$ Adaptation to target domain}}}\\
        \STATE $\textbf{E}_t \leftarrow$ solution of Eq. (\ref{tar_embedding}). 
        \quad{$//$ \textit{\small \textcolor{gray}{Initialization}}}
        \FOR{$\text{iter} \leftarrow 1 ~\text{to}~n_{epoch}$}
        \STATE $\mathcal{L}_{kt}\leftarrow0$.
        \FOR{$\{I^i, y^i\} \in D_t$}
        \STATE $\{\textbf{M}_q, \textbf{M}_{qr}\} \leftarrow$ 
        solution of Eqs. (\ref{tar_concept1}) and (\ref{tar_concept4}).
        
        \STATE $\hat{y} \leftarrow \text{Dec}(\text{Enc}(\textbf{M}_q, \textbf{M}_{qr}))$.\quad{$//$ \textit{\small \textcolor{gray}{Predict $\hat{y}$}}}
        \STATE $\mathcal{L}_{kt} \leftarrow \mathcal{L}_{kt} + \mathcal{L}_{\text{CE}}(\hat{y}, y^i)$.\quad{$//$ \textit{\small \textcolor{gray}{Calculate the loss}}}
        \ENDFOR
        \STATE Update $\textbf{E}_t$ by minimizing $\mathcal{L}_{kt}$.
        \ENDFOR
    \end{algorithmic}
\end{algorithm}

After learning the domain-generalizable KT model, we delve into transferring the model to a new target domain, which remains unseen during the training phase. Firstly, we randomly initialize the concept embeddings for the target domain. Following this, we design a novel question representation for the target domain that not only preserve the specific concepts of the target domain but also demonstrates adaptability to the \emph{concept prototypes} learned from the source domains, thereby preventing overfitting.

\textbf{Initialization for target concept embeddings.} We begin by initializing the concept embeddings for the target domain, namely target concept embeddings. Given that concepts in the target domain remains unseen, a feasible solution for initializing the target concept embeddings is to randomly select ones from the \emph{concept prototypes} to form the matrix $\textbf{E}_t$ of target concept embeddings, which can be formulated as:
\begin{equation}
    \textbf{E}_t = [e_{t1}, e_{t2}, ..., e_{t{n_c}}],\label{tar_embedding}
\end{equation}
where $e_{ti}$ represents the $i$-th target concept embedding ($1\leq i\leq n_{c} $), and $n_{c}$ is the number of concepts in target domain. 

\textbf{Question representation of target domain.} We aim to effectively leverage the \emph{concept prototypes} learned from various source domains to represent the question embeddings of the target domain. Firstly, as in Eq. (\ref{embedding_source}), we calculate the concept-based question embedding $e_{q_t}$ as follows:
\begin{equation}
        e_{q_t} = \frac{\sum_{i} \mathbb{I}(\textbf{Q}_{iq_t} = 1) e_{ti}}{\sum_{i} \mathbb{I}(\textbf{Q}_{iq_t} = 1)}, \label{tar_concept2}
\end{equation}
where $e_{ti}$ is the concept embedding retrieved from $\textbf{E}_t$. 
Then, given the potential risk of overfitting due to limited student interactions in the target domain, we further constrain the question embedding to move closer to its nearest \emph{concept prototype}. Thus, in the target domain, the question embedding $e_{target}$ can finally be  calculated as:
\begin{align}\label{tar_concept1}
    &j = \text{argmin}_{1 \leq i \leq k} \, ||e_{q_t} - e_{c_i}||, \\
    &e_{target} = (1-\lambda) e_{c_j} + \lambda e_{q_t}, \label{tar_concept4}
\end{align}
where $e_{c_j}$ represents the nearest \emph{concept prototype} to $e_{q_t}$, $\lambda$ is a hyperparameter used to control the degree of influence that the target concept embedding has over the final question embedding. This ensures that the question embedding $e_{target}$ not only captures the relevant features of the target domain but also remains aligned with the generalized knowledge encoded in the \emph{concept prototypes}, reducing the likelihood of overfitting.

\textbf{Adaptation of target concept embeddings.} Our KT model is finetuned on the target domain by utilizing the question embedding $e_{target}$ within the cross-entropy loss. During this phase, optimization is confined to the target concept embedding $\textbf{E}_t$, with all the other parameters including $\theta_{enc}$ and $\theta_{dec}$, respectively, remaining fixed. 
In this way, our model will effectively adapt to the target domain while leveraging the \emph{concept prototypes} embeddings learned during concept aggregation. For further clarity, the whole process of concept aggregation detailed in \textbf{Algorithm \ref{alg1}}.

In concept aggregation, the space complexity of our feature embedding module is expressed as \( O\left(\sum_i^n n_{c_i} + n_{c_t} + k\right) \), where \( n_{c_i} \) denotes the number of concepts in the source domain \( i \), \( n_{c_t} \) denotes the number of concepts in the target domain, and \( k \) is the number of \emph{concept prototypes} involved in concept aggregation.

\subsection{Relation-Aware Knowledge Encoder}

Although the proposed DGKT framework enables existing knowledge tracing (KT) models to achieve domain generalizability and enhances their performance in new domains, these models are not inherently designed for cross-domain tasks. A significant challenge impeding the effective transfer of these models to new domains is the limited availability of training data, which hampers the model's ability to fully exploit the information embedded in students' interaction sequences.

To mitigate this limitation, we propose a domain-generalizable relation-aware knowledge encoder (RA-Encoder), explicitly designed to address the challenge of multi-source domain adaptation in knowledge tracing. This encoder leverages an attention mechanism to capture the relationships between questions and concepts across different timesteps. In this section, we provide a detailed exposition of the RA-Encoder, including its architecture and the DGrKT model utilizing RA-Encoder.

\begin{wrapfigure}{r}{0.5\textwidth} 
    \centering
    \includegraphics[width=\linewidth]{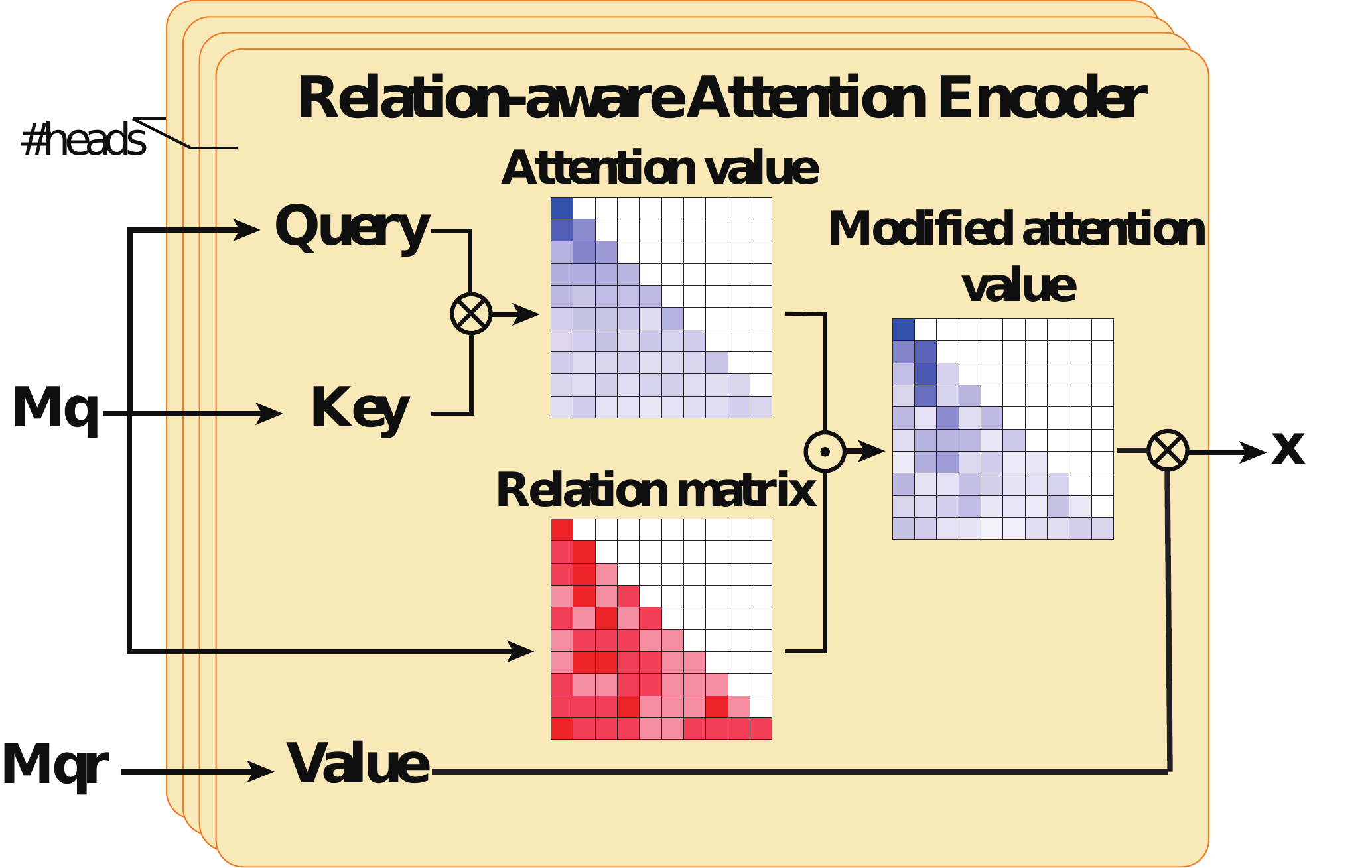}
     \caption{The architecture of relation-aware attention. Relation-aware attention focus more on related exercises and concepts.}
    \label{relation-based}
\end{wrapfigure}

\textbf{Relation-Aware Attention Encoder}
In the Relation-Aware Attention Encoder (RA-Encoder), the attention value $\alpha$ is modulated according to the relation matrix between any two distinct timesteps, as illustrated in Fig. \ref{relation-based}. Our underlying assumption is that the more similar a question a student previously answered is to the current question, the more valuable the information from that previous time becomes. Therefore, we have categorized the relationships between different timesteps into three distinct types to facilitate the learning process of the relation-aware attention encoder.

Given two timesteps \( i \) and \( j \), and their corresponding interactions \( (q_i, r_i) \) and \( (q_j, r_j) \), the relevance \( R(i, j) \) between two questions \( q_i \) and \( q_j \) can be represented as:
\begin{equation}
    R(i,j) = \mathbb{I}(q_i=q_j) + \mathbb{I}(C_i \cap C_j \neq \emptyset),
\end{equation}
where \( C_i \) and \( C_j \) are the concept sets associated with questions \( q_i \) and \( q_j \), respectively, and $C_i \cap C_j \neq \emptyset$ implies the presence of shared concepts. Thus, \( R(i, j) = 0 \) indicates that \( q_i \) and \( q_j \) are irrelevant, \( R(i, j) = 1 \) indicates that the two questions may be relevant due to their association with shared concepts, and \( R(i, j) = 2 \) indicates \( q_i \) and \( q_j \) are the same question. 

Following the procedure of attention mechanism, we employ $e_q$ to corresponds to query ($Q$) and key ($K$), and $e_{qr}$ corresponds to value ($V$). In the attention block, the attention value $\alpha_{ij}$ is calculated as:
\begin{equation}
\begin{split}
    Q_i &= W_qe_{q_i}, \\
    K_j &= W_ke_{q_j}, \\
    \alpha_{ij} &= \text{Softmax}(\frac{Q_i^TK_j}{\sqrt{d}}),
\end{split}
\end{equation}
where $W_q$ and $W_k \in \mathbb{R}^{d \times d}$ are the linear transformation matrices, and $d$ is the dimension of features.

Typically, within the attention mechanism, the attention value \( \alpha \) is directly used to compute the weighted sum. In contrast, the relation-aware attention encoder in our model seeks to explicitly distinguish among various relationships. Consequently, we modify the attention value \( \alpha_{ij} \) based on the relevance \( R(i,j) \). Specifically, for the timestep \( j \), the adjusted attention value \(\boldsymbol{\Tilde{\alpha}}_j\) is calculated as follows:
\begin{equation}
\begin{split}
    \lambda_{ij} &= \begin{cases}
            a & \text{if } R(i,j) = 0 \\
            b & \text{if } R(i,j) = 1 \\
            c & \text{if } R(i,j) = 2 
            \end{cases}, \\
    \boldsymbol{\Tilde{\alpha}}_j
 &= [\lambda_{1j}\alpha_{1j}, \lambda_{2j}\alpha_{2j}, ..., \lambda_{(j-1)j}\alpha_{(j-1)j}], \\
\end{split}
\end{equation}
where \( \lambda_{ij} \) signifies the relevance weight, and $a,b$ and $c$ is the learned parameters. Let \(0 < a < b < c \), this condition ensures that \( \lambda_{ij} \) increases with the relevance \( R(i,j) \). Subsequently, the weighted sum can be computed using the adjusted attention value as follows:
\begin{equation}
\begin{split}
    V_i &= W_v e_{qr_i}, \\
    x_j &= \sum_{i=1}^{j-1} \frac{\lambda_{ij}\alpha_{ij}}{\sum \boldsymbol{\Tilde{\alpha}}_j} V_i,
\end{split}
\end{equation}
where \( W_v \in \mathbb{R}^{d \times d} \) is the linear matrix for the value vectors \( V_i \), and \( x_j \) represents the output of the relation-aware attention encoder for the timestep $j$.

By replacing the knowledge state encoder with the RA-Encoder in the DGKT framework, we introduce the \textbf{Domain-Generalizable Relation-Aware Knowledge Tracing model (DGrKT)}.

Given a question embedding matrix \(\textbf{M}_q\) and a question-response embedding matrix \(\textbf{M}_{qr}\), the outputted knowledge state can be represented as:

\begin{equation}
\begin{split}
    x_t &= \text{RA-Encoder}(\mathcal{Q}=\textbf{M}_q, \mathcal{K}=\textbf{M}_q, \mathcal{V}=\textbf{M}_{qr}), \\
    h_t &= \text{SeqIN}(x_t),
\end{split}
\end{equation}

where \(\mathcal{Q}\), \(\mathcal{K}\), and \(\mathcal{V}\) represent the query, key, and value, respectively, in the attention mechanism. $\text{SeqIN}(\cdot)$ is the aforementioned sequence instance normalization.
As discussed earlier, the knowledge state decoder can then transform 
$h_t$ into the probability of a student answering the next question correctly. Consistent with our DGKT, we apply concept aggregation to facilitate generalization to the target domain.

\section{Experiments}

In this section, we evaluate the performance of our proposed DGKT framework and DGrKT model using five well-known knowledge tracing benchmark datasets on multi-source domain adaptation knowledge tracing. Also, we conduct visualization experiments along with cluster analysis that manifest the effectiveness of our proposed concept aggregation.

\subsection{Datasets and Setting}
We evaluate the performance of our DGKT framework and DGrKT using five well-known KT benchmark datasets: ASSISTment 2009\footnote{https://sites.google.com/site/assistmentsdata/home/assistment-2009-2010-data/skill-builder-data-2009-2010}, ASSISTment 2015\footnote{https://sites.google.com/site/assistmentsdata/home/2015-assistments-skill-builder-data}, ASSISTment 2017\footnote{https://sites.google.com/view/assistmentsdatamining/dataset}, Algebra 2005\footnote{https://pslcdatashop.web.cmu.edu/KDDCup/}, and Junyi\footnote{http://www.junyiacademy.org/}. The first three datasets are provided by an online education system called ASSISTment~\cite{feng2009addressing} with different exercises and students, which are widely used to evaluate KT models. Junyi~\cite{Junyi} was collected from an E-learning platform called Junyi Academy, which was established in 2012. Algebra 2005 is provided by the KDDcup 2010 Educational Data Mining challenge containing 13–14 year old students’ interaction on Algebra questions. Each dataset is composed of extensive student interaction records collected from its respective educational platform. Based on these records, we constructed a personalized sequence of question-response interactions for each student, which were then segmented into fixed-length sequences of 200 interactions. To ensure data quality, students with fewer than 20 total interactions were excluded. The final dataset was partitioned into training and testing sets using an 80:20 split. For convenience, these datasets are abbreviated as ASSIST09, ASSIST15, ASSIST17, ALGEBRA and Junyi in the following.

Table~\ref{tab:dataset_stats} summarizes the statistics of these datasets. The number of students ranges from 565 (ALGEBRA05) to 10,451 (ASSIST15), and the number of unique problems varies significantly, with ALGEBRA05 containing the most (211,397) and ASSIST15 containing the least (100). The total number of student attempts also varies widely, with Junyi having the largest volume (4,371,160 attempts), while ASSIST09 has the least (330,116 attempts). 
These datasets present different levels of data sparsity and distribution shifts, making them ideal benchmarks for evaluating the cross-domain capability of our proposed method.

In the DGKT framework based model, after conducting a grid search, the dimension $d$ of feature embeddings is determined to be 256, and the number $k$ of concept clusters is set to 5. The optimization of the model employs the Adam optimizer with a learning rate of 0.0001, and the batch size is fixed at 32.

Our approach adopts a multi-source domain adaptation setup, where one dataset serves as the target domain, while the remaining four datasets act as four source domains. For concept feature learning, we conduct 12,000 epochs of training on the source domains. In each epoch, every source domain is sampled for a batch and the model is trained on four batches of data. Subsequently, for concept prototypes refinement, we similarly conduct 6,000 epochs of training.

On the target domain, we test our method with different scales of training data, from a single batch of data to 8 batches of data, and a fine-tuning process of 50 epochs is conducted. This approach allows us to assess the effectiveness of knowledge tracing across all five datasets.

\begin{table}[t]
    \centering
    \footnotesize
    \caption{Statistics of the five benchmark datasets used in our experiments.}
    \label{tab:dataset_stats}
    \begin{tabular}{lcccc}
        \toprule
        Dataset & \# Students & \# Problems & \# Concepts & \# Attempts \\
        \midrule
        ASSIST09  & 2314  & 26521  & 124  & 330116  \\
        ASSIST15  & 10451 & 100    & 100  & 624487  \\
        ASSIST17  & 1705  & 3162   & 102  & 942771  \\
        ALGEBRA05 & 565   & 211397 & 113  & 813569  \\
        Junyi     & 10000 & 707    & 40   & 4371160 \\
        \bottomrule
    \end{tabular}
\end{table}

\subsection{Comparison Results and Analysis}
We present a comparative analysis between our proposed DGKT framework, the DGrKT model, and five prominent knowledge tracing (KT) models, \emph{i.e.,} DKT~\cite{piech2015deep}, SAINT~\cite{choi2020towards}, AKT~\cite{ghosh2020context}, SimpleKT (abbreviated a s SKT)\cite{liu2023simplekt}, and RobustKT (abbreviated as RBKT)\cite{guo2025enhancing}.
DG-DKT, DG-SAINT, DG-AKT, DG-SKT, and DG-RKT are instantiated from our DGKT framework, corresponding to the five KT models above.
\textbf{For fair comparison, we initially pre-train these KT models on all available source domains, followed by fine-tuning them on the specific target domain.} Methods including DKT, AKT, SimpleKT and RobustKT are reimplemented by their public source codes, while SAINT is reimplemented based on their papers. Also, the hyperparameters are set as the values provided by the literature.

\begin{table*}
    \centering
    \renewcommand{\arraystretch}{1.1}
    \setlength{\tabcolsep}{2.5pt}
    \scriptsize 
    \caption{The AUC results of four different knowledge tracing methods on five target domain datasets with only a few interactions in the target domain and sufficient interaction sequences in the other four source domains.}
    \label{tab1}
    \begin{tabular}{l|c|cc|cc|cc|cc|cc|c}
        \hline
        Datasets  & batch & DKT & DG-DKT & SAINT & DG-SAINT & AKT & DG-AKT & SKT & DG-SKT  & RBKT & DG-RBKT & DGrKT \\
        \hline
        \multirow{4}{*}{ASSIST09} & 1 & 0.683 & 0.721 & 0.706 & 0.718 & 0.704 & 0.718 & 0.706 & 0.713 & 0.695 & 0.706 & \textbf{0.729} \\
                                  & 2 & 0.685 & 0.727 & 0.711 & 0.727 & 0.713 & 0.723 & 0.713 & 0.715 & 0.704 & 0.715 & \textbf{0.737}\\
                                  & 4 & 0.687 & 0.732 & 0.716 & 0.731 & 0.716 & 0.729 & 0.716 & 0.720 & 0.711 & 0.720 & \textbf{0.740}\\
                                  & 8 & 0.701 & 0.734 & 0.719 & 0.735 & 0.720 & 0.736 & 0.720 & 0.721 & 0.719 & 0.729 & \textbf{0.743}\\
        \hline
        \multirow{4}{*}{ASSIST15} & 1 & 0.658 & 0.694 & 0.680 & 0.683 & 0.672 & 0.690 & 0.670 & 0.691 & 0.668 & 0.675 & \textbf{0.717}\\
                                  & 2 & 0.671 & 0.705 & 0.691 & 0.699 & 0.682 & 0.701 & 0.684 & 0.692 & 0.674 & 0.685 & \textbf{0.729}\\
                                  & 4 & 0.675 & 0.710 & 0.696 & 0.708 & 0.687 & 0.705 & 0.691 & 0.695 & 0.695 & 0.701 & \textbf{0.732}\\
                                  & 8 & 0.677 & 0.716 & 0.702 & 0.714 & 0.698 & 0.719 & 0.694 & 0.696 & 0.705 & 0.717 & \textbf{0.732}\\
        \hline        
        \multirow{4}{*}{ASSIST17} & 1 & 0.600 & 0.675 & 0.636 & 0.649 & 0.640 & 0.661 & 0.636 & 0.655 & 0.627 & 0.641 & \textbf{0.677}\\
                                  & 2 & 0.607 & 0.681 & 0.647 & 0.654 & 0.656 & 0.673 & 0.650 & 0.664 & 0.632 & 0.660 & \textbf{0.684}\\
                                  & 4 & 0.614 & 0.685 & 0.655 & 0.661 & 0.656 & 0.683 & 0.661 & 0.669 & 0.661 & 0.681 & \textbf{0.689}\\
                                  & 8 & 0.617 & 0.693 & 0.661 & 0.667 & 0.664 & 0.685 & 0.669 & 0.674 & 0.678 & 0.691 & \textbf{0.696}\\
        \hline        
        \multirow{4}{*}{ALGEBRA}   & 1 & 0.693 & 0.749 & 0.751 & 0.765 & 0.739 & 0.756 & 0.741 & 0.748 & 0.688 & 0.717 & \textbf{0.775}\\
                                   & 2 & 0.709 & 0.760 & 0.756 & 0.772 & 0.755 & 0.775 & 0.752 & 0.757 & 0.717 & 0.749 & \textbf{0.781}\\
                                   & 4 & 0.722 & 0.764 & 0.762 & 0.779 & 0.759 & 0.781 & 0.761 & 0.769 & 0.768 & 0.771 & \textbf{0.784}\\
                                   & 8 & 0.732 & 0.776 & 0.765 & 0.785 & 0.765 & 0.790 & 0.764 & 0.775 & 0.777 & 0.784 & \textbf{0.795}\\
        \hline
        \multirow{4}{*}{Junyi} & 1 & 0.607 & 0.678 & 0.688 & 0.699 & 0.673 & 0.679 & 0.678 & 0.685 & 0.664 & 0.662 & \textbf{0.724}\\
                               & 2 & 0.637 & 0.684 & 0.689 & 0.701 & 0.688 & 0.692 & 0.686 & 0.688 & 0.678 & 0.685 & \textbf{0.727}\\
                               & 4 & 0.659 & 0.689 & 0.692 & 0.708 & 0.692 & 0.702 & 0.691 & 0.699 & 0.687 & 0.691 & \textbf{0.728}\\
                               & 8 & 0.662 & 0.691 & 0.692 & 0.711 & 0.697 & 0.709 & 0.695 & 0.702 & 0.694 & 0.696 & \textbf{0.730}\\
        \hline
    \end{tabular}
\end{table*}

\textbf{Multi-source Domain Adaptation for Cold-start Knowledge Tracing Tasks.} The AUC results for all compared KT methods in the multi-source domain adaptation are presented in Table \ref{tab1}.These results highlight the effectiveness of the DGKT framework in enhancing knowledge tracing performance with limited training data. In the target domain, DGKT framework demonstrates significant improvements due to its robust generalization capabilities. All knowledge tracing models integrated with the DGKT framework exhibit superior performance compared to traditional knowledge tracing models pretrained on source domains. Moreover, our DGrKT performs the best due to its specially designed attention mechanism. As the size of the data increases, both our DGKT framework and DGrKT consistently maintains its advantage. 

\begin{figure}[H]
    \centering
    \begin{minipage}[t]{0.48\textwidth}
        \centering
        \includegraphics[width=\linewidth]{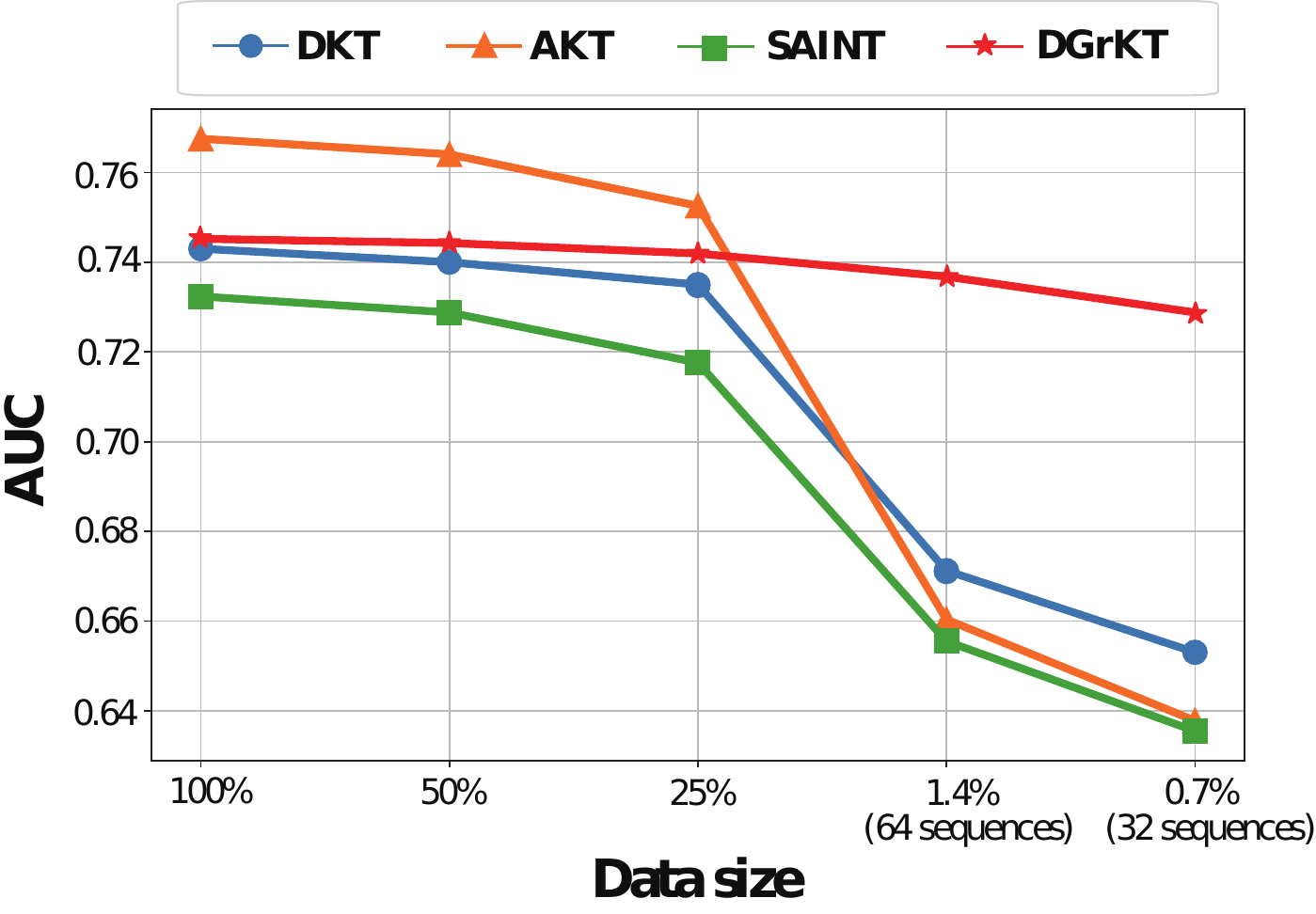}
        \caption{The AUC result for three existing KT models (DKT \cite{piech2015deep}, AKT \cite{ghosh2020context}, SANIT \cite{choi2020towards}) on ASSISTment 2009 with different data sizes compared to DGrKT.}
        \label{data_plot}
    \end{minipage}
    \hfill
    \begin{minipage}[t]{0.48\textwidth}
        \centering
        \includegraphics[width=\linewidth]{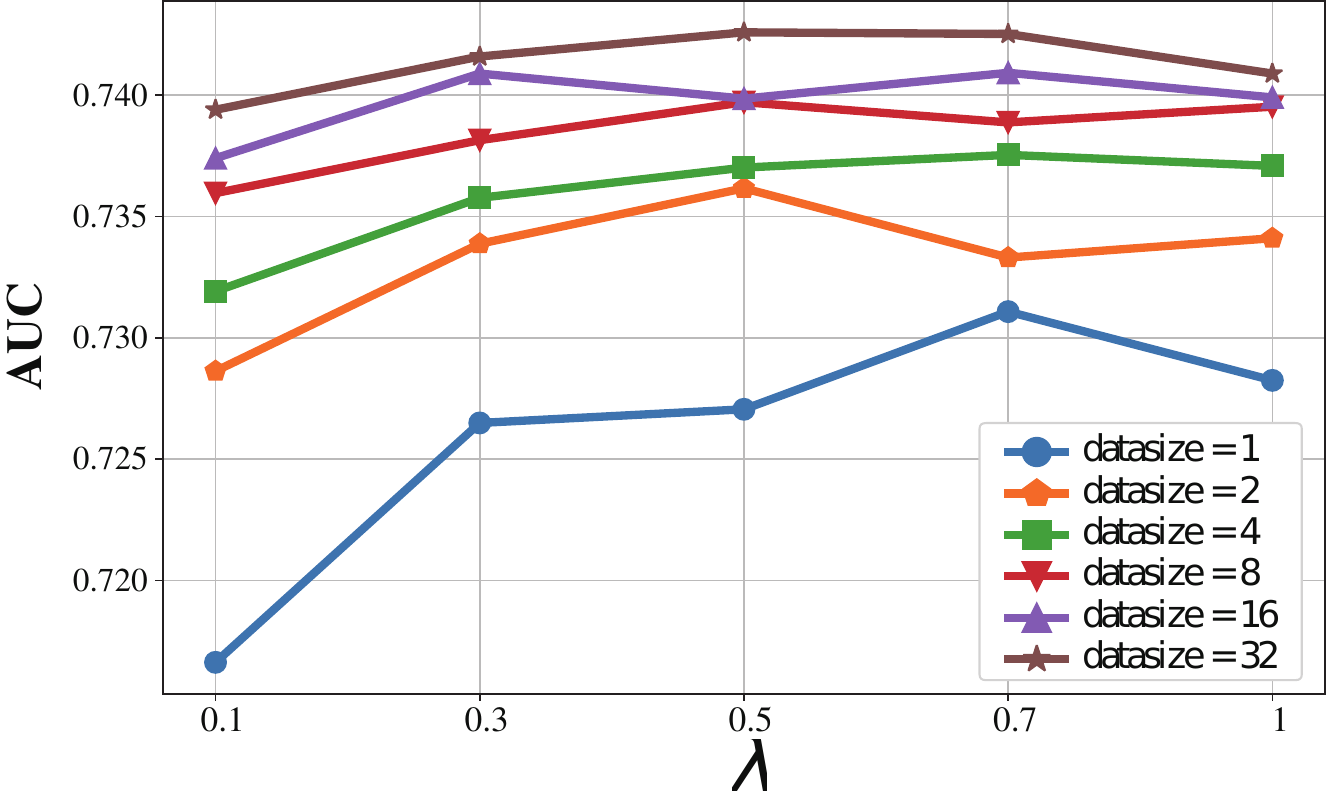}
\caption{The AUC result for different $\lambda$ with different data sizes. Each colored line represents the AUC as $\lambda$ changes under a specific data size setting.}
        \label{lam_fig}
    \end{minipage}
\end{figure}

Moreover, we compare our DGrKT with existing KT models with different data size, as shown in Fig. \ref{data_plot}. We find that as data size decreasing, other KT models exhibit a significant drop in performance. However, the performance of DGrKT declines at a much slower rate, indicating that it is highly adaptable to limited data size.

\subsection{Ablation Studies.}

\begin{table}
    \centering
    \renewcommand{\arraystretch}{1.0}
    \setlength{\tabcolsep}{5pt}
    \footnotesize 
    \caption{The AUC results for different varints of DGrKT on five target domain datasets.}
    \label{ab1}
    \begin{tabular}{l|c|ccccc}
        \hline
        Models  & batch num & ASSIST09 & ASSIST15 & ASSIST17 & ALGEBRA & Junyi \\
        \hline
        \multirow{4}{*}{\makecell{DGrKT \\ -w/o CFL}} & 1 & 0.699 & 0.670 & 0.624 & 0.701 & 0.691 \\
                                                      & 2 & 0.705 & 0.689 & 0.649 & 0.756 & 0.704 \\
                                                      & 4 & 0.716 & 0.707 & 0.667 & 0.764 & 0.709 \\
                                                      & 8 & 0.726 & 0.719 & 0.675 & 0.772 & 0.713 \\
        \hline
        \multirow{4}{*}{\makecell{DGrKT \\ -w/o CR}}  & 1 & 0.711 & 0.697 & 0.646 & 0.750 & 0.700 \\
                                                      & 2 & 0.717 & 0.709 & 0.663 & 0.762 & 0.706 \\
                                                      & 4 & 0.725 & 0.718 & 0.672 & 0.767 & 0.712 \\
                                                      & 8 & 0.728 & 0.723 & 0.681 & 0.774 & 0.714 \\
        \hline        
        \multirow{4}{*}{\makecell{DGrKT \\ -w/o TA}}  & 1 & 0.709 & 0.695 & 0.645 & 0.752 & 0.697 \\
                                                      & 2 & 0.718 & 0.712 & 0.662 & 0.763 & 0.707 \\
                                                      & 4 & 0.724 & 0.717 & 0.672 & 0.767 & 0.711 \\
                                                      & 8 & 0.729 & 0.723 & 0.680 & 0.780 & 0.713 \\
        \hline        
\multirow{4}{*}{DGrKT}    & 1 & \textbf{0.729} & \textbf{0.717} & \textbf{0.677} & \textbf{0.775} & \textbf{0.724}   \\
                                   & 2 & \textbf{0.737} & \textbf{0.729} & \textbf{0.684} & \textbf{0.781} & \textbf{0.727} \\
                                   & 4 & \textbf{0.740} & \textbf{0.732} & \textbf{0.689} & \textbf{0.784} & \textbf{0.728} \\
                                   & 8 & \textbf{0.743} & \textbf{0.732} & \textbf{0.696} & \textbf{0.795} & \textbf{0.730} \\
        \hline
    \end{tabular}
\end{table}

\textbf{Concept Aggregation.}
We conducted ablation experiments to investigate the impact of our proposed concept aggregation methods, which include concept feature learning (CFL), concept prototypes refinement (CR), and target concept embedding adaptation (TA). Specifically, we compare our DGrKT model with three variants: DGrKT-w/o CFL, DGrKT-w/o CR, and DGrKT-w/o TA. The details of these variants are presented below:

\begin{itemize}
\item \textbf{DGrKT-w/o CFL:} This variant omits concept feature learning. The model is directly trained on the target domain without pretraining on the source domain.
\item \textbf{DGrKT-w/o CR:} This variant omits concept prototypes refinement. After concept feature learning, the model is directly fine-tuned on the target domain without concept prototypes refinement.
\item \textbf{DGrKT-w/o TA:} This variant omits target concept embedding adaptation. The model is fine-tuned on the target domain using randomly initialized embeddings, and question embeddings are calculated by Eqn. (\ref{tar_concept2}).
\end{itemize}

From Table \ref{ab1}, we can observe that DGrKT without Concept Feature Learning (CFL) shows a significant decrease in AUC. This indicates that CFL is a crucial procedure, as it derives meta-knowledge from several source domains. Additionally, concept aggregation proves to be beneficial for learning a robust concept embedding for DGrKT, as well as for adapting the target concept embeddings effectively.

\begin{table}
    \centering
    \renewcommand{\arraystretch}{1.1}
    \footnotesize 
    \caption{The AUC results for different varints of DG-DKT on five target domain datasets.}
    \label{ab2}
    \begin{tabular}{c|c|cccccc}
        \hline
        Models  & batch num & ASSIST09 & ASSIST15 & ASSIST17 & ALGEBRA & Junyi \\
        \hline
        \multirow{4}{*}{-}        & 1 & 0.712 & 0.689 & 0.657 & 0.715 & 0.634 \\
                                  & 2 & 0.717 & 0.700 & 0.677 & 0.743 & 0.653  \\
                                  & 4 & 0.726 & 0.704 & 0.680 & 0.764 & 0.681  \\
                                  & 8 & 0.731 & 0.706 & 0.690 & 0.765 & 0.682  \\
        \hline
        \multirow{4}{*}{BN}       & 1 & 0.718 & 0.691 & 0.671 & 0.748 & 0.668  \\
                                  & 2 & 0.722 & 0.702 & 0.679 & 0.759 & 0.669  \\
                                  & 4 & 0.730 & 0.710 & 0.691 & 0.762 & 0.680  \\
                                  & 8 & 0.729 & 0.715 & 0.697 & 0.775 & 0.675  \\
        \hline        
        \multirow{4}{*}{LN}       & 1 & 0.713 & 0.688 & 0.651 & 0.746 & 0.749  \\
                                  & 2 & 0.725 & 0.704 & 0.671 & 0.752 & 0.760  \\
                                  & 4 & 0.727 & 0.713 & 0.679 & 0.760 & 0.764  \\
                                  & 8 & 0.730 & 0.719 & 0.695 & 0.772 & 0.776  \\
        \hline        
\multirow{4}{*}{SeqIN}     & 1 & \textbf{0.721} & \textbf{0.694} & \textbf{0.679} & \textbf{0.749} & \textbf{0.678}  \\
                           & 2 & \textbf{0.727} & \textbf{0.705} & \textbf{0.689} & \textbf{0.760} & \textbf{0.684}  \\
                           & 4 & \textbf{0.732} & \textbf{0.710} & \textbf{0.695} & \textbf{0.764} & \textbf{0.689}  \\
                           & 8 & \textbf{0.734} & \textbf{0.716} & \textbf{0.703} & \textbf{0.776} & \textbf{0.691}  \\
        \hline
    \end{tabular}
\end{table}

\textbf{Sequence Instance Normalization.}
Moreover, we conduct ablation experiment on Sequence Instance Normalization (SeqIN). Thus, we compare DG-DKT with one variant without any normalization method and two variants with BN and LN respectively. 

We omit instance normalization because its calculation inevitably includes all timesteps, leading to information leakage in knowledge tracing. The results, shown in Table \ref{ab2}, demonstrate the advantage of SeqIN. Specifically, DG-DKT without SeqIN (DG-DKT -w/o SeqIN) shows a significant decrease in AUC, indicating that the normalization method is crucial in such cross-domain tasks and Sequence Instance Normalization is a better normalization module for multi-source domain adaptation knowledge tracing.

\subsection{Visualization}
We demonstrate the effectiveness of our proposed method via visualization methods, including visualizing feature embedding via $t$-SNE, student's mastery and attention score of DGrKT.

\begin{figure*}[htbp]
  \centering
  \captionsetup{font=scriptsize}

  \begin{minipage}[b]{0.23\textwidth}
    \centering
    {\small
        \[
        \mathcal{A}\text{-distance}=1.82
        \]
    }
    \includegraphics[width=\textwidth]{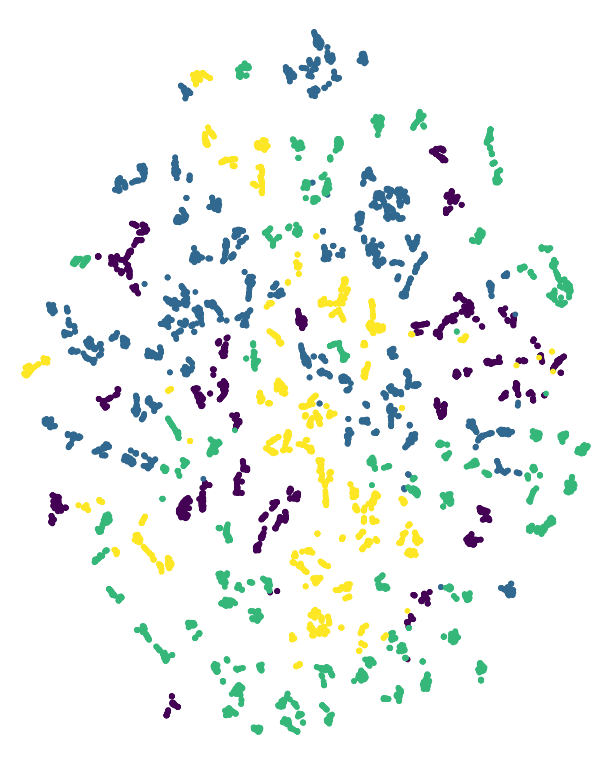}
    \caption*{\scriptsize (a) No normalization}
    \label{fig:subfig1}
  \end{minipage}
  \hfill
  \begin{minipage}[b]{0.23\textwidth}
    \centering
    {\small
        \[
        \mathcal{A}\text{-distance}=1.23
        \]
    }
    \includegraphics[width=\textwidth]{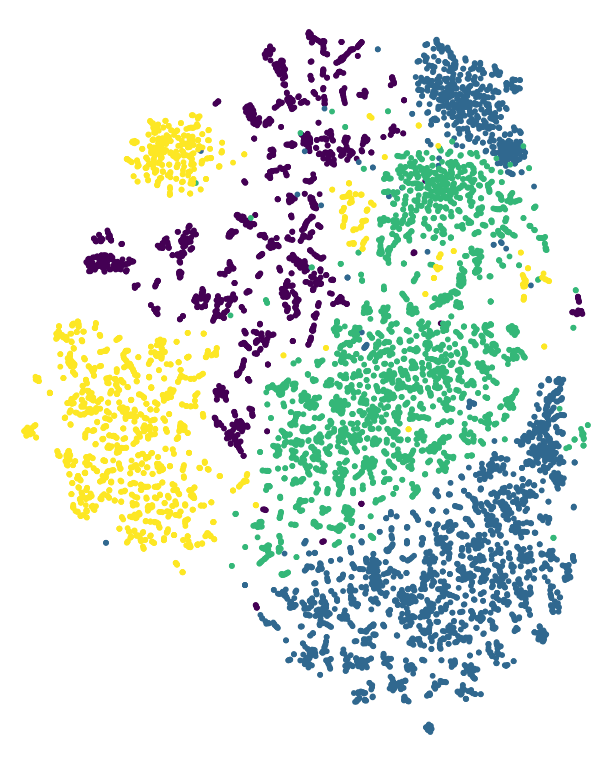}
     \caption*{\scriptsize (b) Batch Normalization}
    \label{fig:subfig2}
  \end{minipage}
  \hfill
  \begin{minipage}[b]{0.23\textwidth}
    \centering
    {\small
        \[
        \mathcal{A}\text{-distance}=1.86
        \]
    }
    \includegraphics[width=\textwidth]{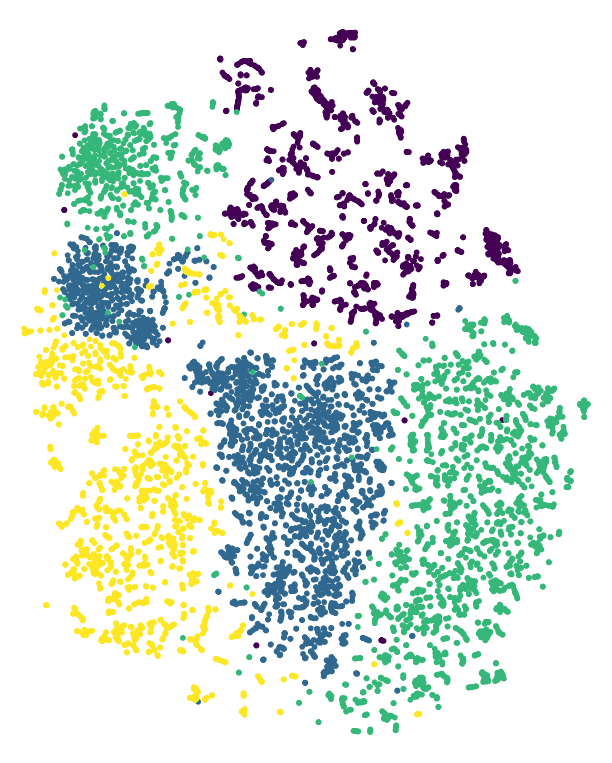}
    \caption*{\scriptsize (c) Layer Normalization}
    \label{fig:subfig3}
  \end{minipage}
  \hfill
  \begin{minipage}[b]{0.23\textwidth}
    \centering
    {\small
        \[
        \mathcal{A}\text{-distance}=0.03
        \]
    }
    \includegraphics[width=\textwidth]{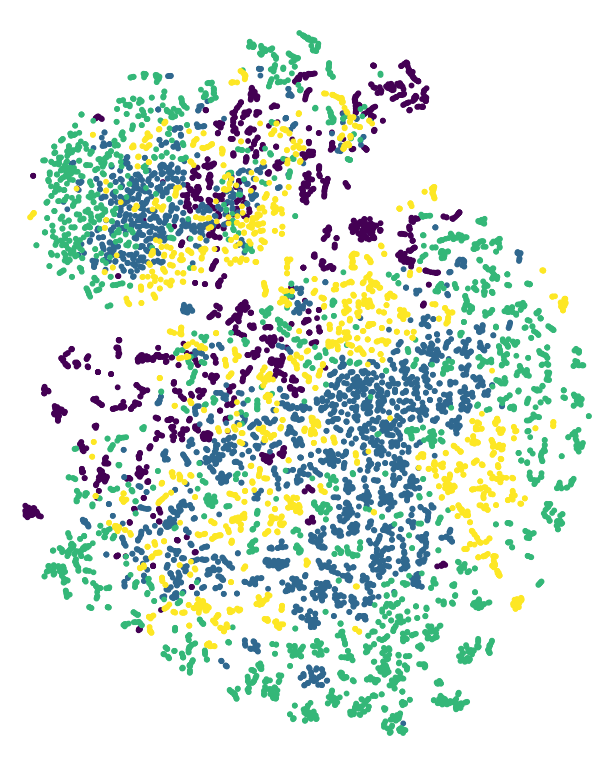}
    \caption*{\scriptsize (d) SeqIN}
    \label{fig:subfig4}
  \end{minipage}

  \caption{\footnotesize The $t$-SNE visualization of feature embedding of student interactions by using different normalization methods. Different colors represent different source domains.}
  \label{tsne}
\end{figure*}

\textbf{Visualization of SeqIN.} To further demonstrate the effectiveness of our SeqIN, we visualize the knowledge state of different domains via $t$-SNE~\cite{van2008visualizing} with different normalization methods in Fig. \ref{tsne}. This figure shows four different source domains (ASSIT15, ASSIT17, Junyi and Algebra05), and each color represents one source domain. Moreover, we use Proxy $\mathcal{A}\text{-distance}$ to measure the discrepancy between different domains \cite{a-dis}.

\begin{figure*}
\vspace{-0.5cm}
\centering
  \includegraphics[width=\columnwidth]{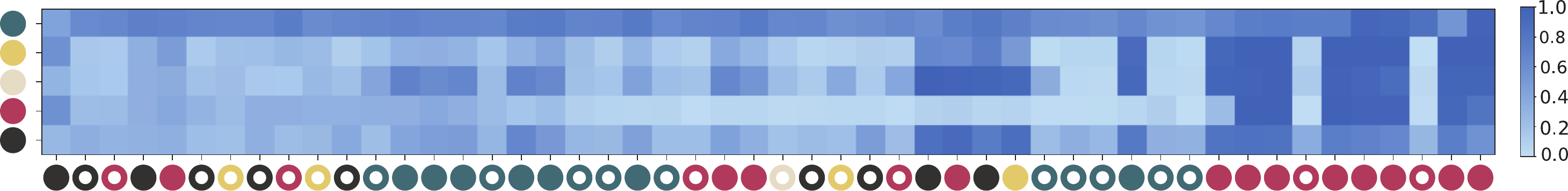}
  \caption{A visualization of a student's mastery across five concept prototypes, based on an interaction randomly selected from ASSIST09. Different colors represent distinct concept prototypes, with hollow circles indicating incorrect responses and filled circles indicating correct responses for the corresponding concept prototype.}\label{vis}
  \label{mastery}
\end{figure*}

From Fig. \ref{tsne} we can observe that the SeqIN can significantly aggregate the knowledge state from different source domains, resulting in the lowest $\mathcal{A}\text{-distance}$, while the BN and LN do not work well on the embedding sequences of student interactions. This shows the efficacy of our SeqIN to reduce the domain discrepancy for knowledge tracing.

\begin{figure}
\centering
\vspace{-0.5cm}
\includegraphics[width=1.25\columnwidth]{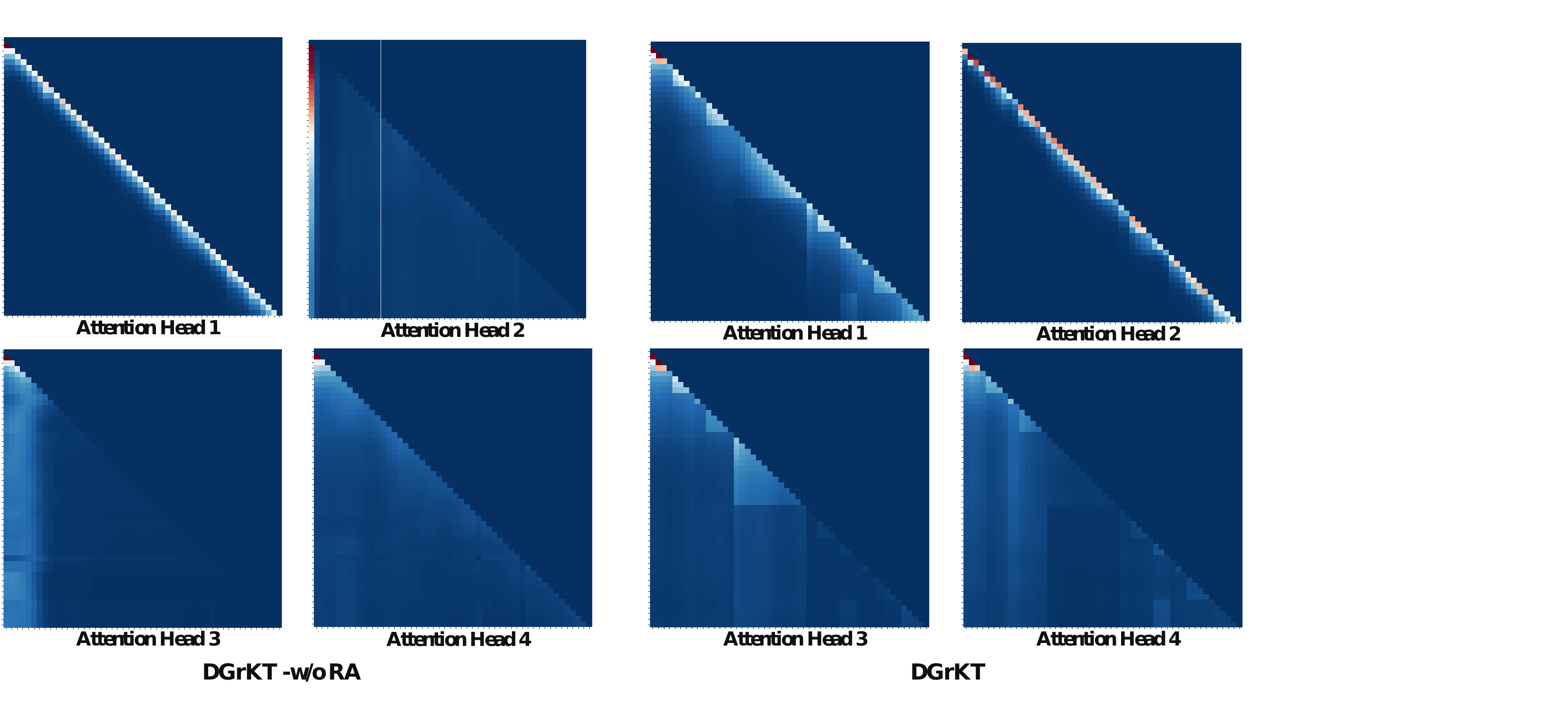}
\caption{Visualization of attention scores from DGrKT and DGrKT-w/o RA. As shown, the attention scores in DGrKT concentrate more effectively on relevant timesteps, while in DGrKT-w/o RA, the attention is spread more uniformly across all timesteps, lacking focus on the crucial related timesteps.}
\label{attentionscore}
\end{figure}

\textbf{Visualization of student's mastery.} We also pivot towards gauging a student's mastery of five concept clusters via visualization, which is calculated by his predicted performance. The interaction sequence is derived from ASSIST09, consisting of the initial 50 interactions, as is shown in Fig. \ref{mastery}. A green rectangle implies that the student correctly answered the corresponding concept cluster while red rectangle means that the student failed. We find that our DGrKT is capable of providing a coarse-grained but effective assessment despite the data scarcity issue.

\textbf{Visualization of attention score.} Here, we demonstrate the effectiveness of relation-aware attention using a heatmap visualization. As shown in Fig. \ref{attentionscore}, we present the attention scores for four attention heads: the left side displays scores from DGrKT without relation-aware attention (DGrKT -w/o RA), while the right side shows scores from DGrKT with relation-aware attention. We observe that with the help of relation-aware attention, DGrKT effectively distinguishes the relationships between different timesteps, focusing on those with closer connections. In contrast, the attention scores in DGrKT -w/o RA are smoother, indicating a reliance primarily on the sequential order.

\textbf{Cluster analysis.} We conducted an analysis on the aggregated concepts from ASSIST09, randomly selecting 8 concepts from each cluster. As shown in Table \ref{cluster}, it is evident that the first cluster primarily consists of fundamental and basic concepts, while the second cluster contains many advanced algebraic concepts. The third, fourth, and last clusters respectively include geometry concepts, advanced data representation, and statistical concepts. Although the concepts are not strictly categorized into five clusters, those with similar patterns in student interactions have been grouped together. These results clearly demonstrate the effectiveness of the concept aggregation process.

\begin{table}[ht]

    \centering
    \footnotesize 
    \caption{Visualization of concept aggregation results.}
    \begin{tabular}{c|p{12cm}}
        \hline
        Cluster & Concepts \\
        \hline
        1 & Scatter Plot, Median, Probability of a Single Event, Area Circle, Ordering Real Numbers, Subtraction Whole Numbers, Multiplication and Division Integers, Order of Operations +,-,/,* ()  \\
        \hline
        2 & Stem and Leaf Plot, Interior Angles Figures with More than 3 Sides, Equivalent Fractions, Square Root, Scientific Notation, Area Irregular Figure, Surface Area Rectangular Prism, Computation with Real Numbers \\
        \hline
        3 & Box and Whisker, Circle Graph, Nets of 3D Figures, Algebraic Solving, Intercept, Slope, Reflection, Volume Cylinder \\
        \hline
        4 & Venn Diagram, Counting Methods, Congruence, Proportion, Prime Number, Algebraic Simplification, Equation Solving More Than Two Steps, Write Linear Equation from Situation \\
        \hline
        5 & Number Line, Mean, Probability of Two Distinct Events, Effect of Changing Dimensions of a Shape Proportionally, Estimation, Scale Factor, Greatest Common Factor, Recognize Linear Pattern \\
        \hline
    \end{tabular}
    \label{cluster}
\end{table}

\subsection{Hyperparameter Analysis}
We conduct several experiments for parameter $\lambda$ (refereed to Eq. \ref{lam_fig}) in concept representation of target domain. We test the AUC of DGrKT with different values of $\lambda$ and different data sizes, shown in Fig. \ref{lam_fig}. When $\lambda$ is set too low, the constraints on the target domain become too strong, preventing the model from learning effective representations of the target domain, which leads to a decrease of AUC in target domain. Conversely, when $\lambda$ is set too high, the constraints diminish, causing the model to lose the information from its learned centroid embedding. As for $\lambda$, it achieves best performance when it is set to 0.7.

\section{Conclusion and Discussion}
This paper introduces a novel approach, Domain Generalizable Knowledge Tracing framework, as a solution to address the data scarcity issue within education systems, which capitalizes on the utilization of multiple source domains to train a versatile KT network, enabling rapid adaptation to new target domains with commendable accuracy. Additionally, we propose the concept prototype to unify concepts from various domains via concept aggregation. Moreover, we propose a domain-generalizable relation-aware knowledge tracing (DGrKT) utilizing relation-aware attention encoder. Experimental evaluations conducted on five benchmark datasets demonstrate substantial improvements when compared to existing KT models.
In conclusion, this study showcases the potential of DGKT in providing a versatile and accurate solution for knowledge tracing across a spectrum of educational domains.

Future research could explore the integration of additional semantic information. The concept aggregation process is purely data-driven, relying solely on student interaction data without integrating richer contextual information. For instance, incorporating textual representations of questions or domain-specific knowledge could potentially enhance generalization across diverse educational settings. Furthermore, beyond knowledge tracing, DGKT could be extended to other applications within educational systems, including personalized learning path recommendation and automatic difficulty adjustment. These extensions would further capitalize on DGKT’s domain-generalizable capabilities, broadening its impact on intelligent education systems.




\bibliographystyle{apalike} 
\bibliography{elsarticle/ref}
\end{document}